\documentclass[a4paper, 10pt, conference]{ieeeconf}      

\IEEEoverridecommandlockouts                              

\makeatletter
\let\NAT@parse\undefined
\makeatother
\usepackage{times}
\usepackage{epsfig}
\usepackage{graphicx}
\usepackage{amsmath}
\usepackage{amssymb}
\usepackage{makecell}
\usepackage{tabularx}
\newcolumntype{L}{>{\raggedright\arraybackslash}X}
\newcolumntype{C}{>{\centering\arraybackslash}X}
\usepackage{array, makecell}
\usepackage{xcolor}
\setcellgapes{3pt}
\usepackage{subfigure}
\usepackage{multirow}

\usepackage[pagebackref=true,breaklinks=true,letterpaper=true,colorlinks,bookmarks=false]{hyperref}

\title{\LARGE \bf
Benefits of temporal information for appearance-based gaze estimation
}

\author{Cristina Palmero$^{1,2}$, Oleg V. Komogortsev$^{3,4}$ and Sachin S. Talathi$^{3}$
\thanks{$^{1}$Universitat de Barcelona, Spain}%
\thanks{$^{2}$Computer Vision Center, Spain}%
\thanks{$^{2}$Facebook Reality Labs, USA}%
\thanks{$^{3}$Texas State University, USA}%
}

\begin{document}

\maketitle
\thispagestyle{plain}
\pagestyle{plain}

\begin{abstract}
State-of-the-art appearance-based gaze estimation methods, usually based on deep learning techniques, mainly rely on static features. However, temporal trace of eye gaze contains useful information for estimating a given gaze point. For example, approaches leveraging sequential eye gaze information when applied to remote or low-resolution image scenarios with off-the-shelf cameras are showing promising results. The magnitude of contribution from temporal gaze trace is yet unclear for higher resolution/frame rate imaging systems, in which more detailed information about an eye is captured. In this paper, we investigate whether temporal sequences of eye images, captured using a high-resolution, high-frame rate head-mounted virtual reality system, can be leveraged to enhance the accuracy of an end-to-end appearance-based deep-learning model for gaze estimation. Performance is compared against a static-only version of the model. Results demonstrate statistically-significant benefits of temporal information, particularly for the vertical component of gaze.
\end{abstract}

\section{Introduction}

Data-driven appearance-based gaze estimation has proven to be a feasible alternative to model-based methods, especially in remote scenarios where lower eye-image resolution does not allow to create a robust model of the eye, and in setups where glints are not available~\cite{hansen2009eye}. During the last decade, deep learning based solutions for gaze estimation have started to emerge due to their excellent performance on a wide range of applications~\cite{lecun2015deep,jiang2019appearance}. Such approaches are usually posed as a regression problem, using Convolutional Neural Networks (CNNs) to extract static features from eye-region~\cite{zhang2015appearance} or whole-face images~\cite{zhang2017s} to estimate the direction of gaze. However, gaze is a dynamic process; depending on the task, we perform different eye movements, e.g., fixations, saccades, smooth pursuit movements, vergence movements, and vestibulo-ocular movements~\cite{leigh2015neurology}. Furthermore, the gaze direction at a certain point in time is strongly correlated to the gaze direction of previous time steps. 

Following this line of reasoning, few works have started to leverage temporal information and eye movement dynamics to increase gaze estimation accuracy with respect to static-based methods. This possibility was first explored by Palmero et al.~\cite{palmerorecurrent}, who proposed to feed the learned static features of all the frames of a sequence to a many-to-one recurrent module to predict the gaze direction of the last sequence frame, improving the state of the art on head-pose independent gaze estimation. Later, Wang et al.~\cite{wang2019neuro} relied on a semi-Markov approach to model the gaze dynamics of fixations, saccades and smooth pursuit movements; per-frame gaze estimates were first computed using a CNN and then refined using the learned dynamic information.  Bidirectional recurrent methods have also been introduced~\cite{kellnhofer2019gaze360, zhou2019learning}, although their applicability is reduced to offline methods as they rely on past and future information. Despite these initial explorations confirming the benefits of temporal information, these works are based on low-to-mid resolution images and low frame rate capture systems ($\sim$30 fps), which do not allow to accurately capture some of the eye movement dynamics, especially saccades, which are characterized by a very high velocity. Therefore, it is yet unclear how and why temporal information improves gaze estimation accuracy for different eye movements.

In this paper, we investigate whether temporal sequences of eye images, captured at a high frame rate (100 Hz) with a  virtual-reality head mounted display (VR-HMD) mounted with two synchronized eye facing cameras, can be leveraged for gaze estimation. Furthermore, we evaluate which eye movements benefit more from the additional temporal information. We specifically focus on fixations (stable point of gaze) and saccades (rapid movement between fixations), two of the most prominent eye movements~\cite{komogortsev2010standardization}. We compare the results obtained with a spatio-temporal model based on a many-to-one CNN-recurrent approach, in contrast to those obtained with a static-only CNN model. This approach was selected as it offers a natural path to addressing our hypothesis in an end-to-end fashion. We evaluate our hypothesis on a newly constructed dataset collected using  the above mentioned VR-HMD, in which 84 subjects of varied appearance were recorded performing a stimulus-elicited saccade task in a VR scenario. Results show that leveraging temporal information of eye image sequences for gaze estimation significantly improves accuracy, in particular for the vertical component of gaze. To the best of our knowledge, this paper presents the first study systematically demonstrating the benefits of temporal information for appearance-based gaze estimation using eye image captures with a  high-resolution, high-frame rate camera system, evaluated on different eye movements.

\section{Methodology}
\label{sec:method}
In this section, we describe the proposed methodology to evaluate the benefits of sequential information for appearance-based gaze estimation models.

\subsection{Spatio-temporal gaze estimation}
In this work, the spatio-temporal gaze estimation task is posed as a regression problem based on monocular eye images. First, spatial features are extracted for each frame $I_i$ of a sequence using a static CNN backbone $g$. The sequence of per-frame features is then fed to a many-to-one recurrent module $r$ to learn sequential dependencies. The recurrent module produces a vector of spatio-temporal features, which is used to finally regress the line of gaze of the last frame of the sequence, $\hat{y}_t$, such that $\hat{y}_t = f(r(g(I_{t-s-1}), \dots, g(I_t)))$, where $f$ denotes the regression function, $t$ corresponds to the last frame of the sequence, and $s$ to the number of previous frames considered for the sequence. The line of gaze is expressed by 2D spherical coordinates, representing yaw (horizontal) and pitch (vertical) angles.

\subsection{Network architecture}

\begin{figure}[t!]
	\centering
	\includegraphics[width=\linewidth]{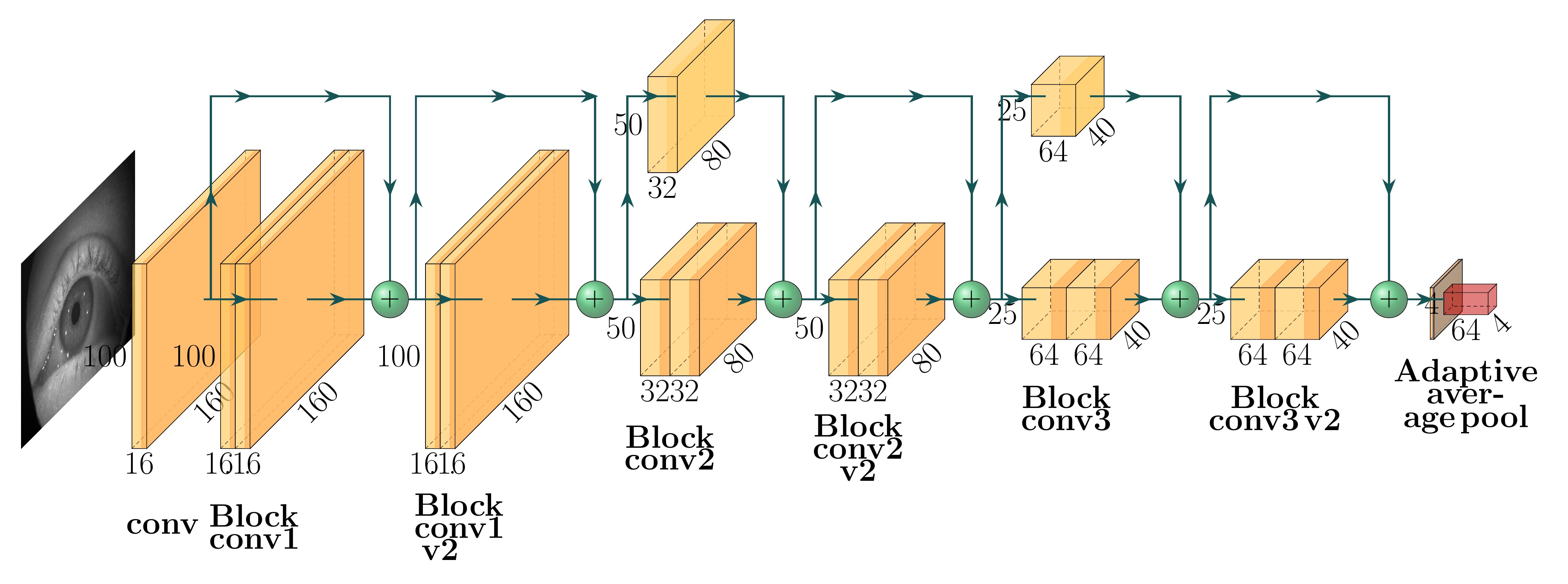}
	\caption{Backbone network used for static gaze estimation, based on a modified ResNet.}
	\label{fig:backbone}
	\vspace{-0.3cm}
\end{figure}

As backbone, we use a modified ResNet architecture ~\cite{he2016deep} with 13 convolutional layers and a final adaptive average pooling layer at the end to decrease the final feature vector size to 64x4x4. The sequence of feature vectors is flattened to serve as input for the recurrent module. The backbone architecture is depicted in Fig. ~\ref{fig:backbone}. 

The recurrent module consists of a single-layer plain Long Short-Term Memory (LSTM) ~\cite{greff2016lstm} with 32 units. The internal memory cell of an LSTM is able to learn long term dependencies of sequential data, avoiding vanishing and exploding gradient problems. The LSTM is unrolled into $s$ time steps, depending on the input sequence length. The output of the recurrent module is further fed to a fully connected (FC) layer with Rectified Linear Unit (ReLU) activation function, which produces a 32-dimensional vector. Finally, a FC layer with linear activation function (i.e. regression) produces the estimated 2D gaze angles.

\subsection{Training strategy}

The network is trained in a stage-wise fashion. First, the static backbone, coupled to a 32-hidden unit FC layer and a 2-hidden unit FC regression layer, is trained end-to-end from scratch on each individual frame of the training data to learn static features. This network is referred to as \textit{Static1} (or \textit{S1}) in Section~\ref{sec:experiments}. Second, the FC layers are discarded, and the recurrent module, coupled to a new 32-hidden unit and 2-hidden unit FC layers, is added to the pre-trained static backbone. The new architecture is further trained end-to-end, fine-tuning the convolutional backbone while training recurrent layer and new FC layers from scratch. By further fine-tuning the backbone weights, the convolutional module is able to learn useful features coming from the sequential information captured by the recurrent module. For this second stage, however, the training data is re-arranged using a sliding window with stride 1, to build input eye image sequences compatible with the many-to-one architecture. Each input sequence is composed of $s$ consecutive frames. This second network is referred to as \textit{S1+LSTM} in Section~\ref{sec:experiments}.

The network was trained using ADAM optimizer~\cite{kingma2014adam}, empirically setting the learning rate to 0.0005, batch size to 32, and weight decay to 0.00001. The learning rate parameter was found to have a large influence on the final accuracy, with lower values not allowing a proper learning of the LSTM. CNN weights were initialized from a uniform distribution. For the LSTM module, input weights were initialized using Xavier uniform, while an orthogonal initialization was used for hidden weights. Biases were set to 0. Early stopping on the validation set was used to select the best model for each training stage, with a maximum number of epochs of 150 for the first stage and 30 for the second. Finally, we used the L1 loss for both training stages, as preliminary experiments showed it to yield slightly lower error than L2 loss.

\section{Experiments}
\label{sec:experiments}
In this section, we describe the experimental setup and evaluate the effectiveness of the spatio-temporal model in comparison to a static-only version for different window lengths and eye movements.

\subsection{Dataset}
\label{sec:dataset}
The study is based on a newly constructed dataset of $100 \times 160$-pixel eye-image sequences captured using a VR-HMD~\footnote{Details on capture setup are beyond the scope of the short paper and will be discussed in a follow-up long form version of the paper. The VR-HMD used was similar to usual commercially available wireless VR-HMDs.}, with two synchronized eye-facing infrared cameras at a frame rate of 100Hz under constant illumination. The dataset consists of 84 subjects with appearance variability in terms of ethnicity, gender, and age, with some of them wearing glasses, contact lenses, and/or make-up. Subjects were recorded while gazing at a moving target on screen. Each recording consisted of a set of patterns with 1s-long randomly-located target fixations at different depths and instantaneous (0.1s) target transitions to elicit saccades. 

\begin{figure}[t!]
	\centering
	\raisebox{-0.5\height}{\includegraphics[width=0.5\linewidth]{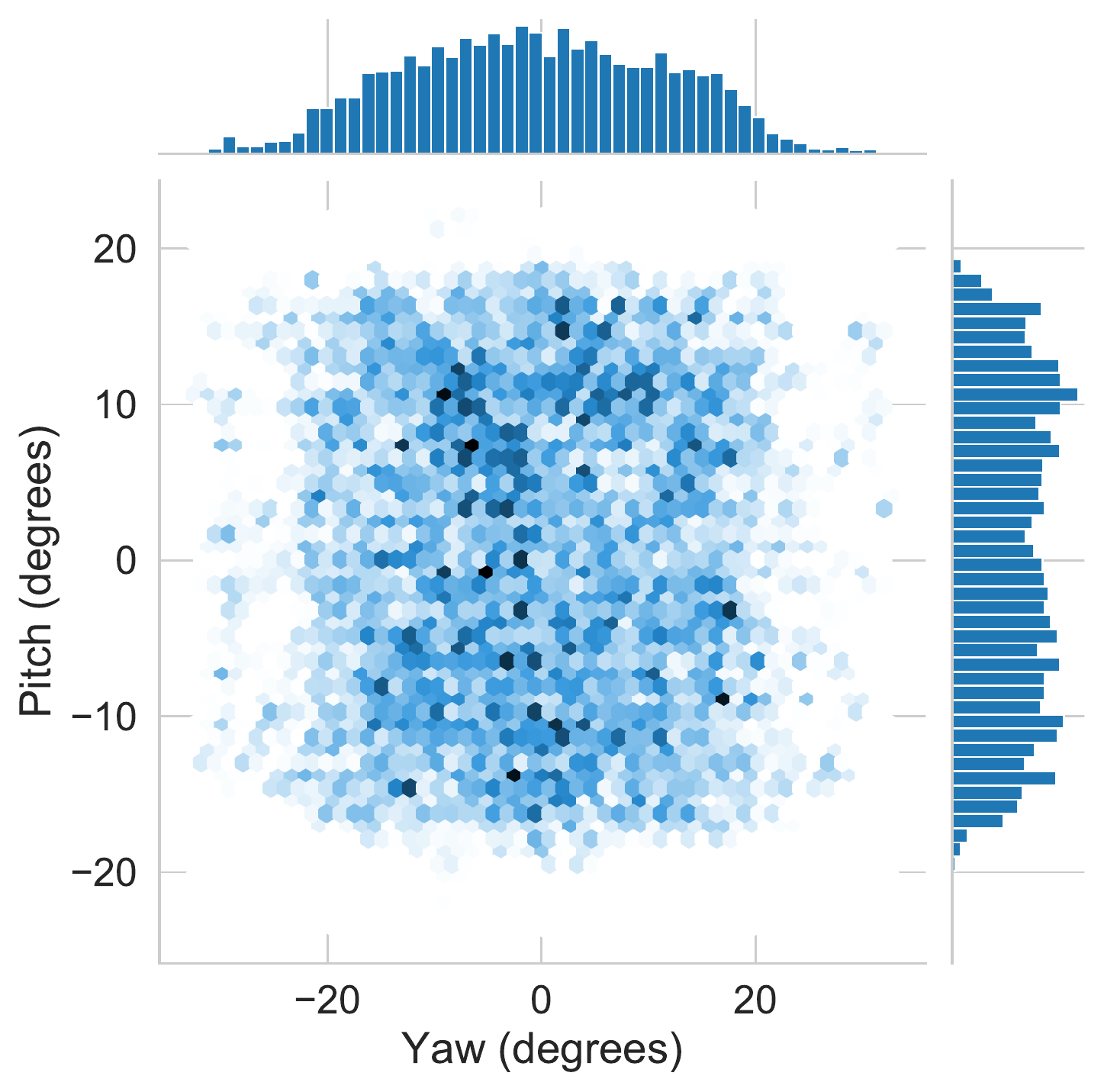}}
	\hfill
	\raisebox{-0.5\height}{\includegraphics[width=0.42\linewidth]{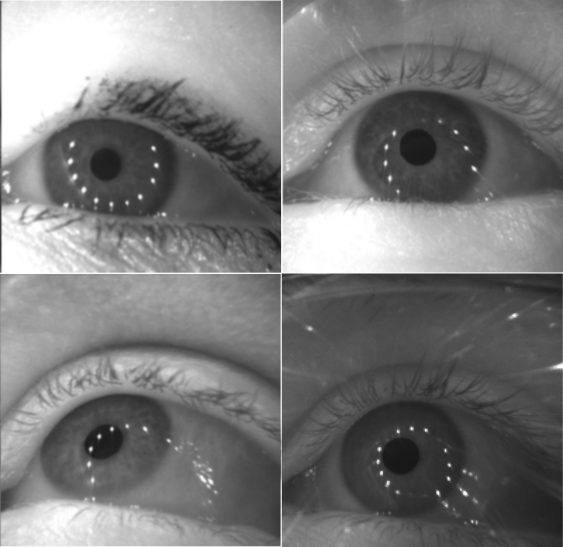}}
	\caption{Gaze distribution (left) and sample eye images (right) from the dataset.}
	\label{fig:dataset_dist}
\end{figure}

Ground truth eye gaze vectors were obtained using a classical user-calibrated glint-based model~\cite{guestrin2006general}. While this approach poses some limitations on the ground truth quality (see Section \ref{sec:limitations}), it still allows us to soundly evaluate our hyphoteses. Frames with no valid gaze vector, or for which the subject was distracted, were discarded, causing most of the recordings to be divided in smaller non-contiguous frame sequences. To keep consistency, the remaining data was further processed by randomly selecting 10 non-overlapping sequences of 100 contiguous frames (1s) each, thus having 1,000 frames per recording and a total of 168,000 eye-region images. Therefore, each sequence can contain fixations only, saccades, or a combination thereof. Fig.~\ref{fig:dataset_dist} shows the gaze angle distribution and sample eye images from the dataset.

\subsection{Experimental protocol}
To perform the experimental evaluation, the processed dataset was partitioned into subject-independent train, validation, and test sets following a 5:1:2.4 ratio. The evaluated models were trained on the training split, using the validation split to optimize model hyper-parameters. Right-eye images and their corresponding ground truth gaze vectors were horizontally flipped to mimic left-eye data. This way, the same model can be used for both eyes, while augmenting the number of data samples.    

Experimental results are reported on the test split, using Mean Absolute Error (MAE) between estimated and ground truth 2D gaze angles as main metric. Due to the sliding window approach followed by the spatio-temporal model, the first frames of a sequence do not obtain a gaze estimation. Therefore, in this section, results are reported on the subset of the test split which does obtain gaze estimates for all evaluated models~\footnote{The longest window length evaluated in this paper is 20 frames; therefore, $81\%$ of the test split is used to report experiment results. Results from models based on smaller window lengths, evaluated on a consequently larger test subset, did not significantly deviate from the results reported herein.}.

\subsection{Addition of temporal information to baseline static model}

First, we use the initial static model (\textit{Static1}) as baseline and compare it to the proposed spatio-temporal model. In particular, we evaluate 4 versions of the spatio-temporal model, each of them trained with different sliding window lengths $s$ in the range $\{5,10,15,20\}$, to assess the effect of the amount of frames used on the final accuracy.

Table ~\ref{tab:sac_res} shows the performance of the evaluated models with respect to each axis individually and simultaneously. We can observe that all spatio-temporal models significantly outperform the static baseline, with up to $19.78\%$ mean error improvement (paired Wilcoxon test, $p<.0001$). While the error for the horizontal axis is higher than the vertical for all models, the addition of temporal information decreases the error by up to $16.91\%$ for the former and $23\%$ for the latter, evidencing that such information is more beneficial for the vertical axis. This is an important contribution with respect to classical or pupil-based methods, as they usually have less accuracy on this axis due to occlusions of the eye limbus caused by eyelids and lashes. With respect to the window length, we can observe that the increase in performance peaks around $s=15$ frames (i.e. 150ms) and then diminishes, indicating that longer-term dependencies are not required to obtain further accuracy gains.

It could be argued that the decrease in error is due to the larger complexity of the spatio-temporal models, as the addition of the LSTM layer highly increases the number of parameters. To validate this possibility, we trained a second static model (\textit{Static2} in Table ~\ref{tab:sac_res}), adding a 128-hidden unit FC layer between the two FC layers from \textit{Static1} model to compensate for the difference in number of parameters between baseline and spatio-temporal models. Results show that, in spite of the smaller number of parameters, even \textit{Static1} outperforms \textit{Static2}, suggesting that \textit{Static2} may be overfitting to the training data. This demonstrates that the increase in complexity is not correlated with a lower error. 

\begin{table}[t!]
	\caption{Mean absolute error (degrees) between ground truth and estimated gaze angles for the different evaluated models, reported on the test set. Standard deviation in brackets.}
	\label{tab:sac_res}
	\begin{tabular}{cccccc}
		\hline
		Method & Window & Yaw & Pitch & Mean \\
		\hline
		Static1 (\textit{S1}) & 1  & 4.02 (4.22) & 3.26 (2.67) & 3.64 (2.59) \\
		Static2               & 1  & 4.26 (4.93) & 3.36 (2.72) & 3.81 (2.93) \\
		S1+LSTM               & 5  & 3.46 (4.03) & 2.57 (2.14) & 3.01 (2.41) \\
		S1+LSTM               & 10 & 3.41 (3.95) & 2.55 (2.14) & 2.98 (2.39)  \\
		S1+LSTM               & 15 & \textbf{3.34 (3.98)} & \textbf{2.51 (2.07)} & \textbf{2.92 (2.37)}  \\
		S1+LSTM               & 20 & 3.39 (3.99) & 2.51 (2.08) & 2.95 (2.39) \\ 
		\hline
	\end{tabular}
\end{table}

\begin{figure}[t!]
	\centering
	\includegraphics[width=\linewidth]{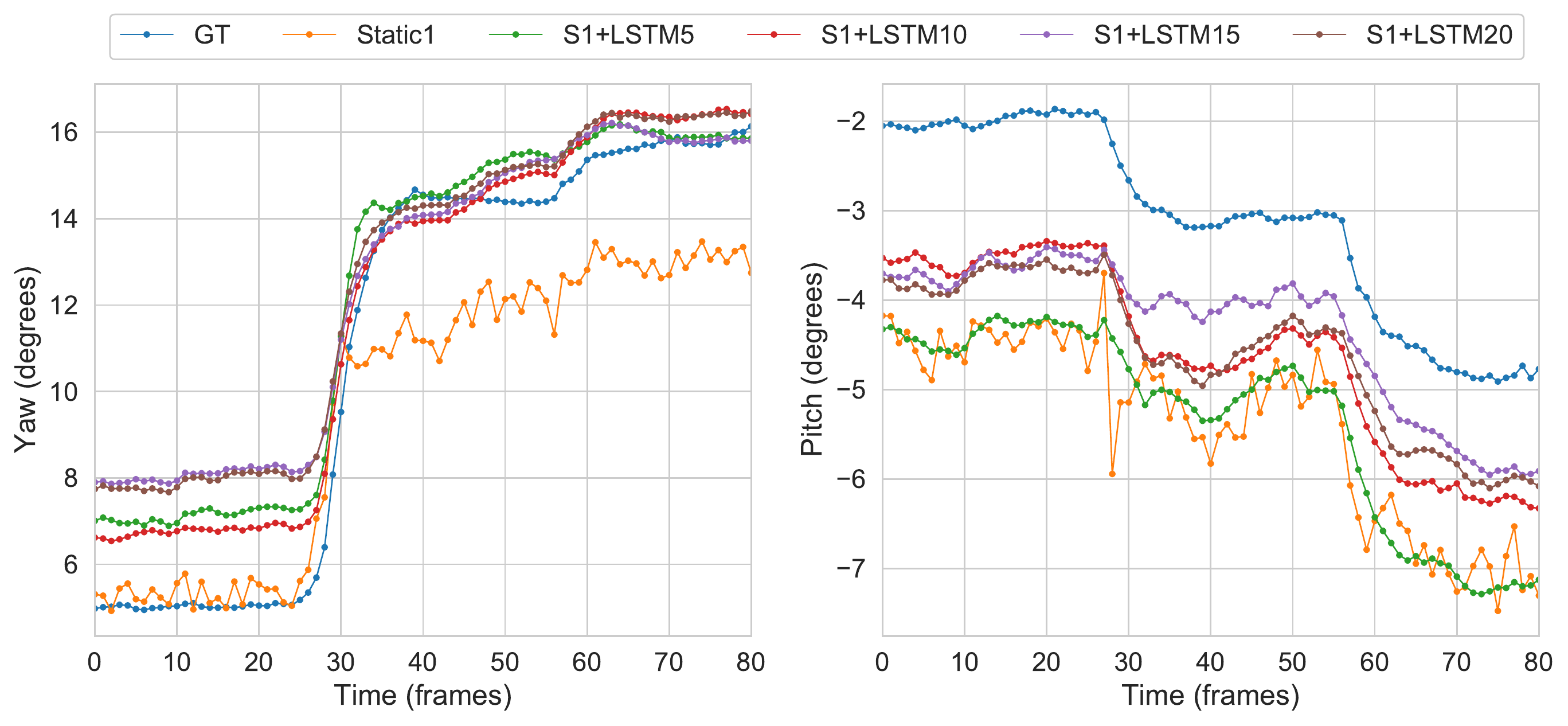}
	\caption{Example of ground truth (GT) and estimated gaze angles for one dataset subsequence.}
	\label{fig:example_sac}
\end{figure}

Fig.~\ref{fig:example_sac} further illustrates the effects of leveraging temporal information, with an example of ground truth and estimated gaze angles during a fixation-saccade transition, for horizontal and vertical axis. We can clearly see the smoothing effect of temporal information as opposed to the noisy static estimation. Furthermore, spatio-temporal estimates are able to more accurately follow the saccade-to-fixation transition. Indeed, using consecutive frames allows the network to better discard noisy features and learn a more useful representation for eye gaze estimation.

\subsection{Contribution of temporal information wrt. eye movement type}
In this subsection, we evaluate the contribution of temporal over static models with respect to different eye movements types. To do so, we use the 20-frame-window spatio-temporal model (\textit{S1+LSTM20}) as the reference temporal model for this experiment. 

\begin{figure}[t!]
	\centering
	\raisebox{0\height}{\includegraphics[width=0.49\linewidth]{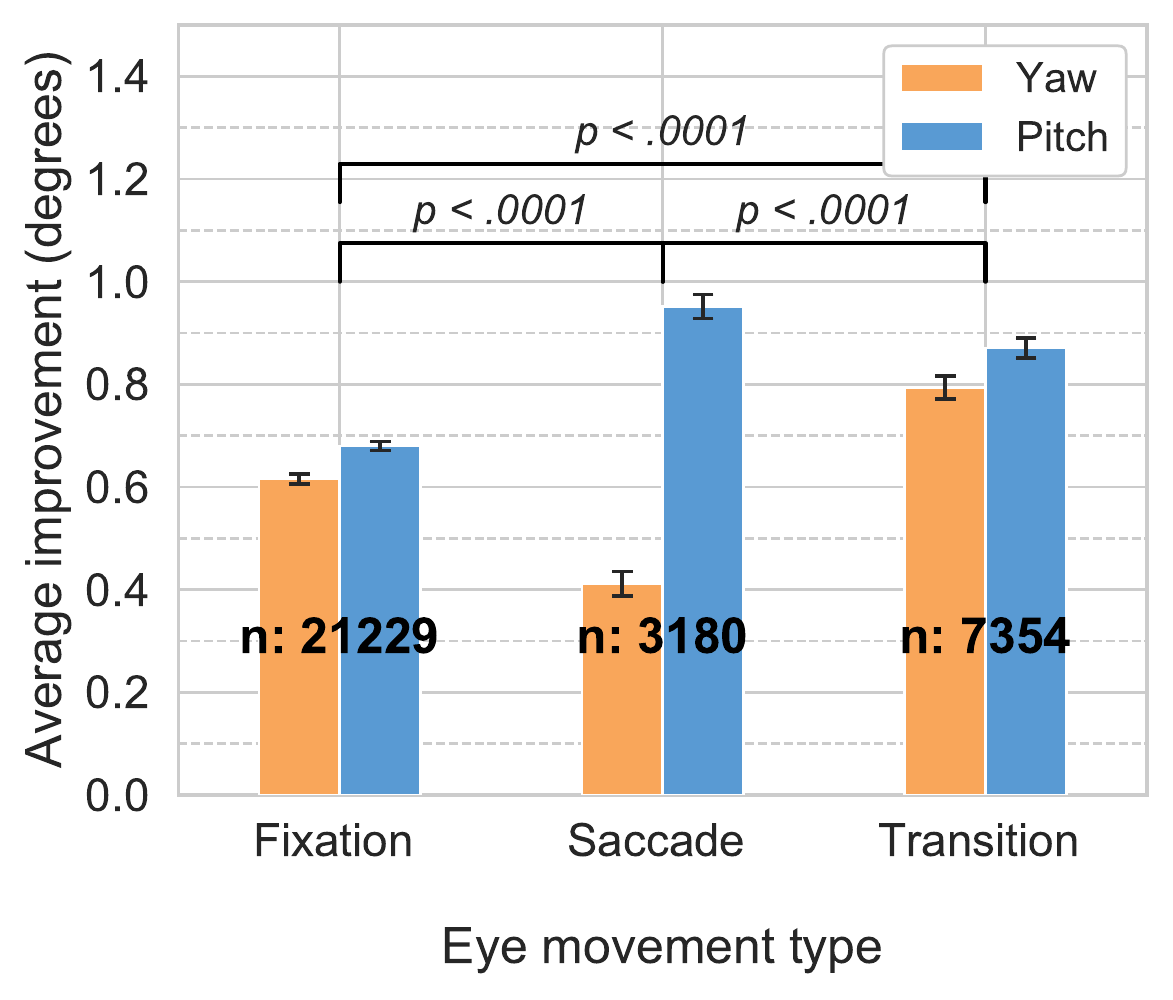}}
	\raisebox{0\height}{\includegraphics[width=0.49\linewidth]{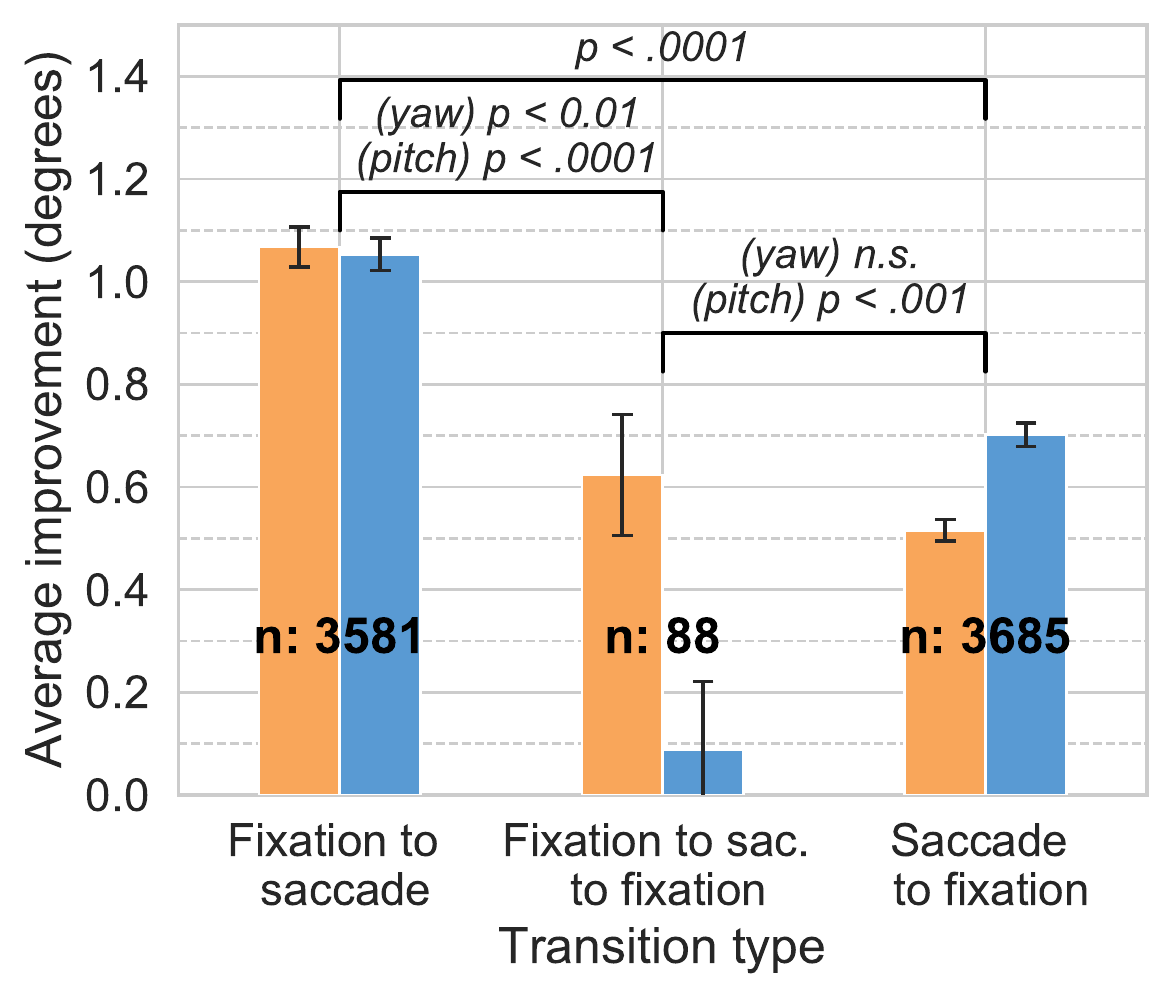}}
	\caption{Average improvement of temporal (\textit{S1+LSTM20}) over static (\textit{Static1}) models per axis (horizontal and vertical), for different eye movement (left) and transition (right) types. Error bars indicate standard error of the mean. Significance computed using the two-sided Kolmogorov-Smirnov test.}
	\label{fig:eye_mov}
\end{figure}

Since our dataset has a mixture of eye movements, to perform this evaluation we manually annotated the test split based on visual inspection of ground truth gaze angles with the following labels per each 20-frame input sequence: \textit{Fixation}, when the eye was virtually static; \textit{Saccade}, if it only included a saccade movement; \textit{Transition}, if it included a combination of fixation and saccades; and \textit{Other}, when the eye status could not be clearly classified. \textit{Transition} sequences were further divided into \textit{Fixation to saccade}, \textit{Saccade to fixation}, or \textit{Fixation to saccade to fixation}, according to their order in time.

Fig.~\ref{fig:eye_mov} depicts the contribution of temporal information with respect to the static baseline for each label and axis. As shown previously, the performance of the vertical axis substantially improves when adding temporal information for fixations and saccades, compared to the horizontal axis. In particular, we observe a substantially higher improvement for saccades, with an average difference of 0.44 degrees between axes. With respect to transitions,  the horizontal axis shows a higher improvement than the vertical axis for fixation-to-saccade transitions. This demonstrates that the spatio-temporal model is indeed taking into account the information coming from the first frames of the input sequence, being more beneficial when the sequence starts with a more stable eye gaze and more correlated eye images to better learn the transition dynamics. As a matter of fact, this improvement is even higher than the one obtained on \textit{Saccade}-only sequences, suggesting that the model is able to learn more representative features of the eye when being presented with a fixation bout first, particularly for the horizontal axis.

\subsection{Effect of appearance}
As a final experiment, we evaluated the differences on reported error for each subject. While the error indeed varied among subjects, we did not find any significant differences with respect to the subjects characteristics (i.e. age, gender, glasses and make-up).

\section{Limitations of the study}
\label{sec:limitations}
This study offers an initial insight as to how and why temporal information benefits gaze estimation when different eye movement types are considered, specifically fixations, saccades, and transitions among them. These movements were elicited using a pattern-based task, which pose a limitation on the directions, velocities and motion trajectories available. Eye movement dynamics are task dependent, thus a more complete study should contain subjects performing further tasks, including natural behaviors. We note that the obtained results are linked to the selected methodology, static backbone and loss used for training. Other backbones would pose different static priors which would affect the final obtained error. Finally, even using state-of-the-art glint-based methods to obtain ground truth eye gaze vectors, this process poses a lower bound on the obtained gaze estimation error, as we are trying to approximate a model to values that can be inherently noisy. This is a limitation present in most of the gaze estimation literature, which evidences the need for better ways to gather accurate ground truth gaze data.

\section{Conclusions}
\label{sec:conclusions}
In this paper, we have analyzed the effect of leveraging sequential information for appearance-based gaze estimation, using previous contiguous image frames along with the current image frame to be estimated. We leverage on deep learning techniques, building a spatio-temporal model consisting of a static CNN network followed by a recurrent module to learn sequential dependencies in eye movements. The dataset consists of high-resolution eye-image sequences, consisting of 84 subjects performing a stimulus-elicited fixation and saccade task within a VR scenario, captured at 100Hz. Results have shown a significant improvement of the spatio-temporal model in comparison to a static-only approach, producing a less noisy estimation. The model is able to learn movement dynamics, with increased accuracy when transitioning from fixation to saccade. In addition, temporal information has demonstrated to be particularly beneficial to improve accuracy on vertical axis estimates.

We hope this study serves as stepping stone for future research on novel methods leveraging time and eye dynamics as additional features. Furthermore, the large differences in performance with respect to gaze axis obtained in this study give rise to consider approaches based on independent models for each gaze component, as opposed to usual jointly-trained methods, to improve final gaze estimation accuracy.


{\small
\bibliographystyle{ieee}
\bibliography{egbib}

\begin{thebibliography}{10}\itemsep=-1pt

\bibitem{greff2016lstm}
K.~Greff, R.~K. Srivastava, J.~Koutn{\'\i}k, B.~R. Steunebrink, and
  J.~Schmidhuber.
\newblock Lstm: A search space odyssey.
\newblock {\em IEEE transactions on neural networks and learning systems},
  28(10):2222--2232, 2016.

\bibitem{guestrin2006general}
E.~D. Guestrin and M.~Eizenman.
\newblock General theory of remote gaze estimation using the pupil center and
  corneal reflections.
\newblock {\em IEEE Transactions on biomedical engineering}, 53(6):1124--1133,
  2006.

\bibitem{hansen2009eye}
D.~W. Hansen and Q.~Ji.
\newblock In the eye of the beholder: A survey of models for eyes and gaze.
\newblock {\em IEEE transactions on pattern analysis and machine intelligence},
  32(3):478--500, 2009.

\bibitem{he2016deep}
K.~He, X.~Zhang, S.~Ren, and J.~Sun.
\newblock Deep residual learning for image recognition.
\newblock In {\em Proceedings of the IEEE conference on computer vision and
  pattern recognition}, pages 770--778, 2016.

\bibitem{jiang2019appearance}
J.~Jiang, X.~Zhou, S.~Chan, and S.~Chen.
\newblock Appearance-based gaze tracking: A brief review.
\newblock In {\em International Conference on Intelligent Robotics and
  Applications}, pages 629--640. Springer, 2019.

\bibitem{kellnhofer2019gaze360}
P.~Kellnhofer, A.~Recasens, S.~Stent, W.~Matusik, and A.~Torralba.
\newblock Gaze360: Physically unconstrained gaze estimation in the wild.
\newblock In {\em Proceedings of the IEEE International Conference on Computer
  Vision}, pages 6912--6921, 2019.

\bibitem{kingma2014adam}
D.~P. Kingma and J.~Ba.
\newblock Adam: A method for stochastic optimization.
\newblock {\em arXiv preprint arXiv:1412.6980}, 2014.

\bibitem{komogortsev2010standardization}
O.~V. Komogortsev, D.~V. Gobert, S.~Jayarathna, S.~M. Gowda, et~al.
\newblock Standardization of automated analyses of oculomotor fixation and
  saccadic behaviors.
\newblock {\em IEEE Transactions on Biomedical Engineering}, 57(11):2635--2645,
  2010.

\bibitem{lecun2015deep}
Y.~LeCun, Y.~Bengio, and G.~Hinton.
\newblock Deep learning.
\newblock {\em nature}, 521(7553):436--444, 2015.

\bibitem{leigh2015neurology}
R.~J. Leigh and D.~S. Zee.
\newblock {\em The neurology of eye movements}.
\newblock OUP USA, 2015.

\bibitem{palmerorecurrent}
C.~Palmero, J.~Selva, M.~A. Bagheri, and S.~Escalera.
\newblock Recurrent cnn for 3d gaze estimation using appearance and shape cues.
\newblock In {\em Proceedings of British Machine Vision Conference (BMVC)},
  2018.

\bibitem{wang2019neuro}
K.~Wang, H.~Su, and Q.~Ji.
\newblock Neuro-inspired eye tracking with eye movement dynamics.
\newblock In {\em Proceedings of the IEEE Conference on Computer Vision and
  Pattern Recognition}, pages 9831--9840, 2019.

\bibitem{zhang2015appearance}
X.~Zhang, Y.~Sugano, M.~Fritz, and A.~Bulling.
\newblock Appearance-based gaze estimation in the wild.
\newblock In {\em Proceedings of the IEEE conference on computer vision and
  pattern recognition}, pages 4511--4520, 2015.

\bibitem{zhang2017s}
X.~Zhang, Y.~Sugano, M.~Fritz, and A.~Bulling.
\newblock It's written all over your face: Full-face appearance-based gaze
  estimation.
\newblock In {\em Proceedings of the IEEE Conference on Computer Vision and
  Pattern Recognition Workshops}, pages 51--60, 2017.

\bibitem{zhou2019learning}
X.~Zhou, J.~Lin, J.~Jiang, and S.~Chen.
\newblock Learning a 3d gaze estimator with improved itracker combined with
  bidirectional lstm.
\newblock In {\em 2019 IEEE International Conference on Multimedia and Expo
  (ICME)}, pages 850--855. IEEE, 2019.

\end{thebibliography}
}

\clearpage

\end{document}